\documentclass[a4paper, 10pt, conference]{ieeeconf}
\IEEEoverridecommandlockouts
\pdfoutput=1
\newif\ifpdf\ifx\pdfoutput\undefined\pdffalse\else\pdfoutput=1\pdftrue\fi
\usepackage{graphicx, epsfig, subfigure}

\title{\LARGE \bf Multi-Modal Local Sensing and Communication\\ for Collective Underwater Systems}

\author{Serge Kernbach, Tobias Dipper, Donny Sutantyo 
\thanks{Institute of Parallel and Distributed Systems, University of Stuttgart, Universit\"atstr. 38, 70569 Stuttgart, Germany, emails \{Serge.Kernbach, Tobias.Dipper, Donny.Sutantyo\}@ipvs.uni-stuttgart.de, pre-print version, published in Proceedings of the 11th International Conference on Mobile Robots and Competitions, Robotica 2011, Lisbon, pp.96-101}%
}

\begin{document}

\maketitle
\thispagestyle{empty}
\pagestyle{empty}

\begin{abstract}
This paper is devoted to local sensing and communication for collective underwater systems used in networked and swarm modes. It is demonstrated that a specific combination of modal and sub-modal communication, used simultaneously for robot-robot and robot-object detection, can create a dedicated cooperation between multiple AUVs. These technologies, platforms and experiments are shortly described, and allow us to make a conclusion about useful combinations of different signaling approaches for collective underwater systems.
\end{abstract}

\section{Introduction}
\label{sec:intro}

Underwater exploration represents a very important economic, technologic and scientific challenge. This is closely related to Arctic and Antarctic offshore resources, pollution monitoring, general oceanographic data collection and, recently, to underwater actuation~\cite{TinyFish10}. Due to very large underwater areas and high damping properties of water, application of multiple Autonomous Underwater Vehicles (AUVs) in cooperative missions seems very promising~\cite{Bingham02}.

For application of AUVs in networked or swarm mode, there is a number of crucial issues: underwater sensing and communication (S\&C), cooperation and mission control, design of AUV platforms, autonomous behavior and several collective aspects of running multiple AUVs. In this work we concentrate on minimalistic local S\&C~\cite{Kornienko_S05d}, and related coordination strategies, being motivated by the following reasons~\cite{Partan06asurvey}.

In several past and running projects devoted to underwater swarms, such as AquaJelly~\cite{AquaJelly}, Angels~\cite{ANGELS}, CoCoRo~\cite{CoCoRo}, a number of AUV platforms and sensing technologies has been developed. These works indicated two important issues: a successful AUV platform needs a dedicated combination of different S\&C technologies, moreover capabilities of underwater cooperation depends on the level of embodiment \cite{Kornienko_S05e} of on-board S\&C systems. In several cases, even a simple multi-modal signal system leads to advanced cooperation~\cite{Kornienko_S05b} (see e.g. the case of multi-agent cooperation~\cite{Kornienko_S04}).

Since swarm approaches rely primarily on local interactions between AUVs~\cite{Kernbach08}, the paper is devoted to local S\&C systems (unmodulated and modulated IR/blue light, RF and electric field), which can be used for robot-robot/robot-object detection and provide sub-modal information, such as direction and distances, as well as can be used for analog and digital communication~\cite{Kornienko_S05a}. These systems, developed for each of the platforms is described in Sec.~\ref{sec:localCom}, whereas Sec.~\ref{sec:conparison} provides general overview over different S\&C technologies. In Sec.~\ref{sec:experiments} we shortly sketch a few behavioral experiments with these systems and finally in Sec.~\ref{sec:conclusion} conclude about useful combinations of S\&C and their embodiment for collective underwater systems.

\section{Comparison of different S\&C systems}
\label{sec:conparison}

In this section we give a short overview over different state-of-the-art S\&C systems in an underwater environment, see e.g.~\cite{Lanbo08}-\cite{Schill04}. Four systems will be compared:

\begin{itemize}
	\item \emph{Sonar}: Sonic waves travel very well under water and the energy and build-space required for generating and receiving them is very low. This approach is used in so-called acoustic modems~\cite{Akyildiz04}. Drawbacks of this approach are, firstly, the relatively low sound travel speed of roughly 1500~m$/$s (much slower than any other S\&C system), and, secondly, multiple reflections causing essential distortions in the signal.
	
	\item \emph{Radio}: Electromagnetic waves are a standard communication method in air; its application under water creates several problems~\cite{Siegel73}. Due to water connectivity, the attenuation of radio waves depends on the used frequency, which in turn results in the size of antenna. High frequencies ($>$~100~MHz) only need a small antenna ($0.1$~m) while their range is restricted to $2.5$~m. Lower frequencies (100~kHz) have a long range (100~m), but need a large antenna (100~m).
	
	\item \emph{Optical}: Using light as a communication channel can provide a compact size of transmitting equipment and acceptable range~\cite{Schill04}. Due to the color dependent attenuation of light in water, the communication range varies between a few centimeters in IR spectra and increases to over a meter by using blue or green light.
	
	\item \emph{Electric Field}: This is a new communication approach. It bases upon generating and measuring electric fields. The build-size and energy required for this system is very small. Unfortunately the attenuation of electric fields is very high, liming the range of this communication channel to less than 1~m. We will discuss this approach in Sec.~\ref{sec:elect}.
\end{itemize}
Table \ref{tab:ComparisonOfDifferentCommunicationChannels} shows an overview of the discussed approaches.
\begin{table}[tp]
\vspace{+4mm}
	\centering
		\begin{tabular}{p{3.5cm}rrr}
			\hline \hline
			Channel & Attenuation & Antenna Size & Range \\
			\hline
			Sonar (30 kHz) & 0.3 dB/m & 0.1 m & 300 m \\
			\hline
			Radio (100 kHz) & 1 dB/m & 100 m & 100 m \\
			Radio (1 MHz) & 4 dB/m & 10 m & 25 m \\
			Radio (100 MHz) & 40 dB/m & 0.1 m & 2.5 m \\
			\hline
			Optical unmodul. (IR 800 nm) & 10 dB/m & 0.1 m & 0.25 m \\
			Opt. modul. (PCM IR 800 nm) & 10 dB/m & 0.1 m & 0.5 m \\
			Optical modul. (blue 460 nm) & 1 dB/m & 0.1 m & 1.2 m \\
			\hline
			Electric Field (2.5 kHz) & 100 dB/m & 0.1 m & 1 m \\
			\hline \hline			
		\end{tabular}
	\caption{Comparison of different communication channels}
	\label{tab:ComparisonOfDifferentCommunicationChannels}
\vspace{-5mm}
\end{table}
Since the used AUVs operate in a swarm mode~\cite{Kornienko_S04a} (large number of AUVs, full decentralization, utilization of swarm approaches for coordination~\cite{Kernbach08Permis}, application of evolutionary approaches~\cite{Kernbach08online}, \cite{Kernbach09_CEC}), in this paper we concentrate on a local S\&C approaches. The S\&C is defined as local, when the communication range $R_c$ (i.e. communication volume $V_c=4/3\pi R_c^3$) does not overstep the second-next-neighbors at average swarm density $D_{sw}=N/V_{sw}$, where $V_{sw}$ is the volume occupied by AUVs and $N$ is their number. Local communication range $R_c$ can be approximated by
\begin{equation}
\label{eq^1}
R_c=^3\sqrt{\frac{V_{sw}}{N 4/3 \pi}},
\end{equation}
where for $V_{sw}=5m^3$ and $N=20$, $R_c$ is about 0,4m. For the platform size 10-15cm, this results in 3 to 4~times the robot length. Generalizing the AUV size up to 50cm, we assume that $R_c^l$ within 0,5-1,2m are local, whereas $R_c^g$ capable to cover the whole $V_{sw}$, i.e. 3-4m, are global.
In the following sections we consider the developed optical and electric field S\&C approaches which are related to $R_c^l$.

\section{Local S\&R approaches}
\label{sec:localCom}

\subsection{Multi-modal Optical System}

As the first developed approach for combined S\&C within $R_c^l$, we describe a specific bi-modal directional optical system, which underlies cooperative behavior of AquaJelly robots~\cite{AquaJelly}. AquaJelly was a project between Festo AG \& Co. KG (coordinator and founder), Effekt-Technik GmbH, and University of Stuttgart intended to create a swarm ($N=20-30$ robots) of autonomous underwater robots, capable of multi-modal interactions and underwater recharging. Robots have been developed and manufactured within a very short time of 8 months in 2007-2008.
\vspace{-2mm}
\begin{figure}[htp]
\centering
\subfigure{\includegraphics[width=.4\textwidth]{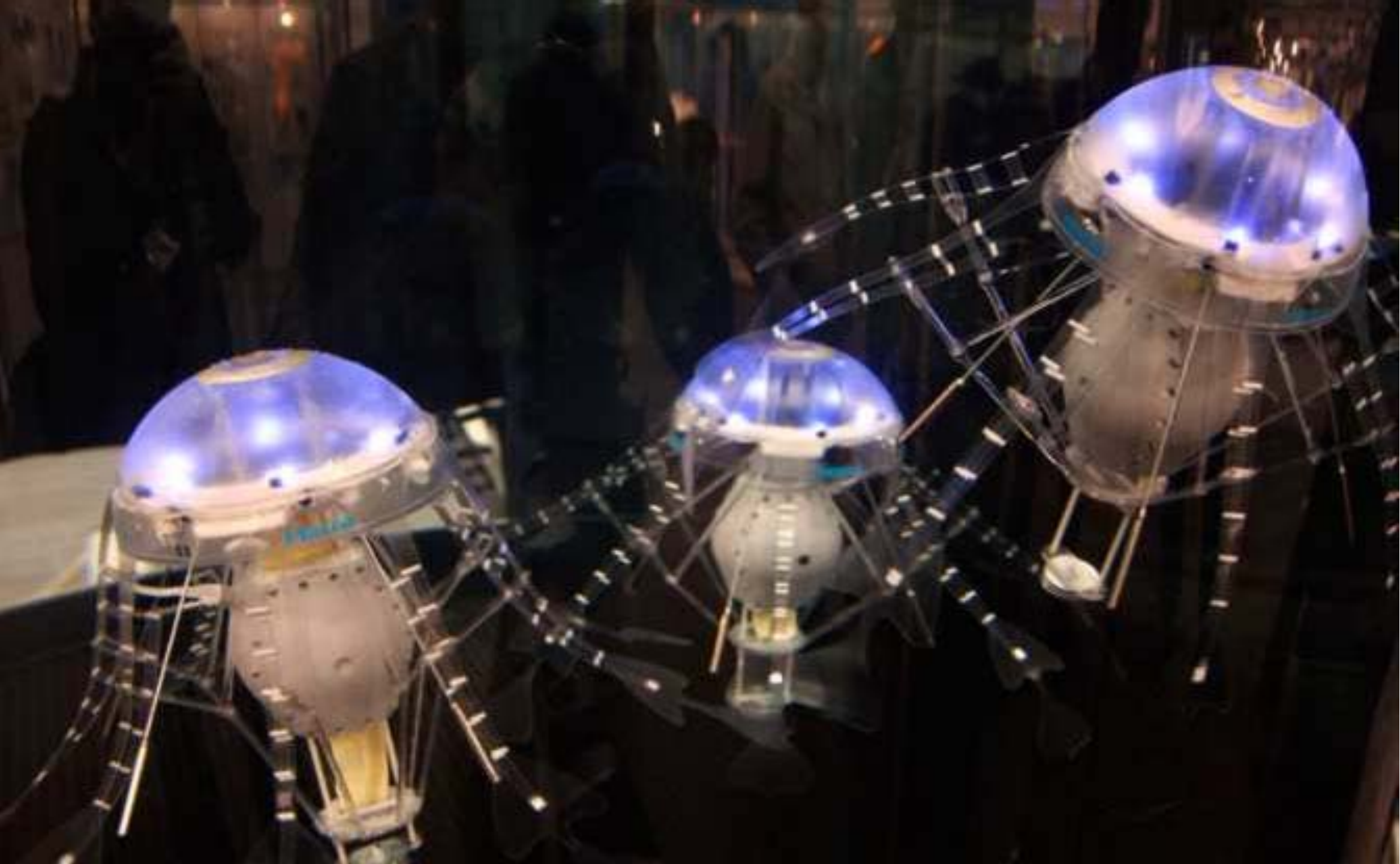}}
\caption{AquaJelly robots.\label{fig:Aquajelly}}
\end{figure}
\vspace{-4mm}

Technical requirements define robot-robot, robot-docking-station, and collisions recognition; cooperative collision avoidance; several types of cooperative behavior around docking station, and vertical movement of robots (robots possess only vertical DoF with a balancing mechanism; this allows an inclined vertical movement). Due to visual effects, which are one of the main developmental goals of this platform, it was decided to use unmodulated blue light. Since several communication approaches should remain invisible for human observers and to make the system more stable to different illumination conditions, it was decided to use additionally 36kHz modulated IR light. The platform possesses also very sensitive pressure and temperature sensors, 3D accelerometer and energy sensor (additional RF system was used for a backup communication with the host). The energy part consisted of 4A/h LiPo accumulator with a power management circuitry and Hot-Swap controller for underwater recharging. Due to low energy consumption, the autonomy lies between several hours and with autonomous recharging is theoretically unlimited.

\begin{figure}[htp]
\centering
\subfigure{\includegraphics[width=.22\textwidth]{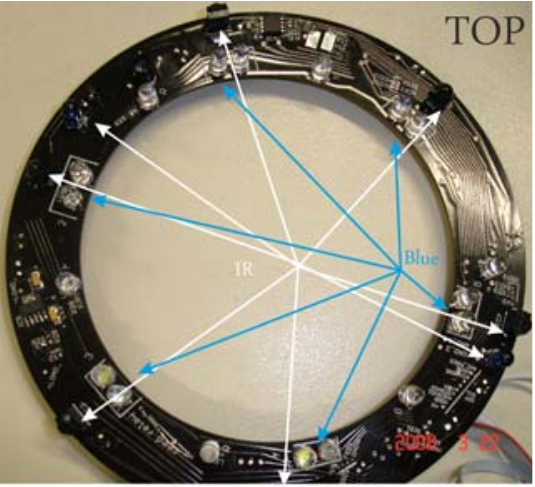}}~
\subfigure{\includegraphics[width=.199\textwidth]{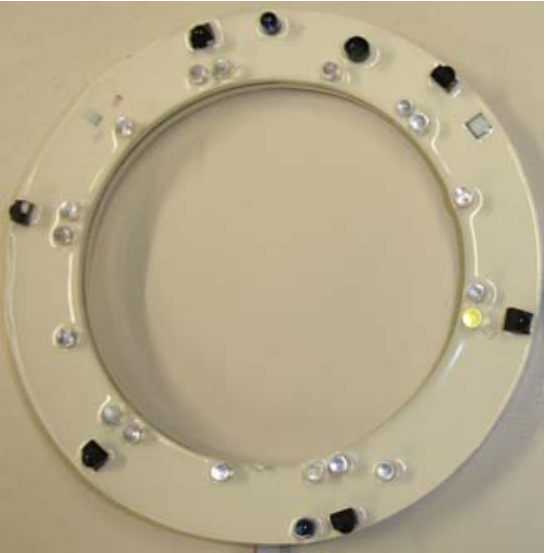}}\\
\vspace{-1mm}
\subfigure{\includegraphics[width=.45\textwidth]{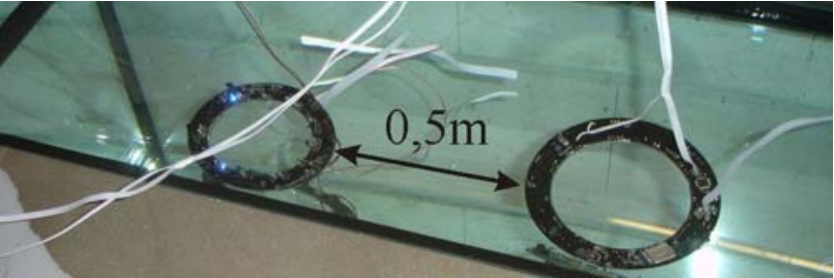}}
\caption{\textbf{(Top-left)} Placement of IR and blue light LEDs in the top of the ring; \textbf{(Top-righ)} Molded ring in white polyurethane; \textbf{(Bottom)} Experimental measurement for IR communication/sensing.\label{fig:Colors}}
\end{figure}

3D omni-directional communication and sensing was one of the main technical requirements. Blue light and IR sensors are used in different ways. Since IR light is more dumped in water, IR channels are very useful for a short range directional communication. Blue light channels are used as unidirectional system, which was mainly used for navigation approaches based on optical pheromone in the improved version of the platform. To enable directional communication, the original platform has 11 IR emitters and receivers with integrated PCM decoder, see Fig.~\ref{fig:Colors}.
\vspace{-2mm}
\vspace{-2mm}

Spatial displacement has an important role, so sidewise IR emitters and receivers are set up each 60 degrees and separated through thick a black-colored PCB on TOP and BOTTOM sides. Three IR sensors are positioned down-side and two up-side. Six blue light LEDs are installed on the top of the platform and through mate cover created almost a homogeneous "light ball" around the robot. All 17 communication channels are independent from each other through an analog multiplexor. To make the platform waterproof, the ring with all sensors was molded in polyurethane. Blue light system has been used in analog mode, whereas IR used a digital PCM-modulated signal. All sensors are directly connected to I/O pins of the Atmel MCU, which can provide around 8-10mA current at 3V, opening angle of all LEDs is about 10 degrees. Communication range of modulated IR is about 0,5m, see Fig.~\ref{fig:Colors}. Due to passive PCM filter, the communication range was fixed on this distance and used in 4kbps communication mode. Average range of the analog blue light system is about 1-1,3m and can be varied by regulating intensity and number of LEDs.

The main idea for using two different optical systems was a split between analog gradient-based interactions between robots (visible for human observer) and digital channels used for robot-objects interactions and for communication, which synchronizes internal states of robots and docking station (invisible for observers). Thus, a combination between analog omni-directional "long-range" and digital directional "short-range" optical systems used in different modifications AquaJelly robots allowed a wide range of different sensing and communication approaches, which result in interesting cooperative behavior of these platforms, see Sec.~\ref{sec:experiments}.

\subsection{Modulated and Encoded Blue Light}
\label{sec:modulatedBlue}

As a further development of the S\&C system, described in the previous section, we intend to use only one light system with modulated blue light for both robot-robot/object detection, distance measurement and digital communication. Digital optical communication is widely used due to its high bandwidth. However, the absence of gradient-based optical guide for sensing and localization makes the digital system less suitable for navigation purposes. Therefore specific protocols are required to extract sub-modal information about distances and orientation from the digital channel. The table~\ref{measurement} shows our underwater measurement results that compare the common modulated IR and blue light communication in many modulation types at the bandwidth of 119kbps.
\begin{table}[htp]
\vspace{-2mm}
\centering
\begin{tabular}{cccc} \hline \hline
Modulation & Transducer & Maximum Communication/Sensing \\
\hline
direct &Infra-red & - / -\\
IrDA & Infra-red & 7 cm / 0-5 cm \\
TV Remote & Infra-red & 5 cm / 0-5 cm \\
QAM & Infra-red & 12 cm / 0-5 cm \\
direct & Blue LED & 20 cm / - / - \\
IrDA & Blue LED & 60 cm / 0-5 cm \\
TV Remote & Blue LED & 45 cm / 3-8 cm \\
QAM & Blue LED & 120 cm / 7-12 cm \\
\hline \hline
\end{tabular}
\caption{Range of underwater optical communication for 119kbps.\label{measurement}}
\vspace{-4mm}
\end{table}

As a digital communication transceiver, the blue light system needs modulator, amplifier, signal conditioner, and protocol encoder/decoder. The one chip solution can be solved by using a CS 8130 IrDA chip from Cirrus Logic. Since the blue light system has a directional S\&C, two channels are not sufficient for the swarm robot to communicate in every direction. The half duplex behavior of each channel makes one channel unable to be applied for sensing, because the sensing mechanism requires to transmit and to receive the sensing signal in one time. Therefore, the position of the transmitter and receiver are swapped with neighboring channels for sensing application.

\begin{figure}[ht]
\centering
\vspace{-2mm}
\subfigure[]{\includegraphics[width=.23\textwidth, height=0.12\textheight]{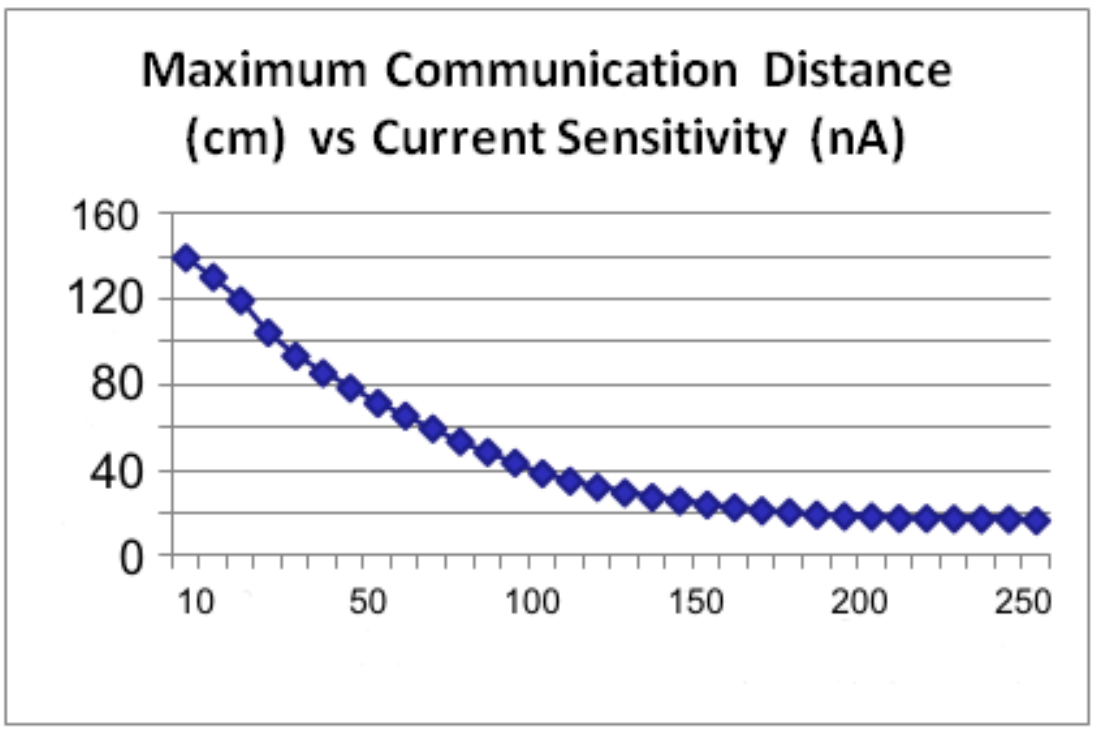}}~
\subfigure[]{\includegraphics[width=.25\textwidth, height=0.12\textheight]{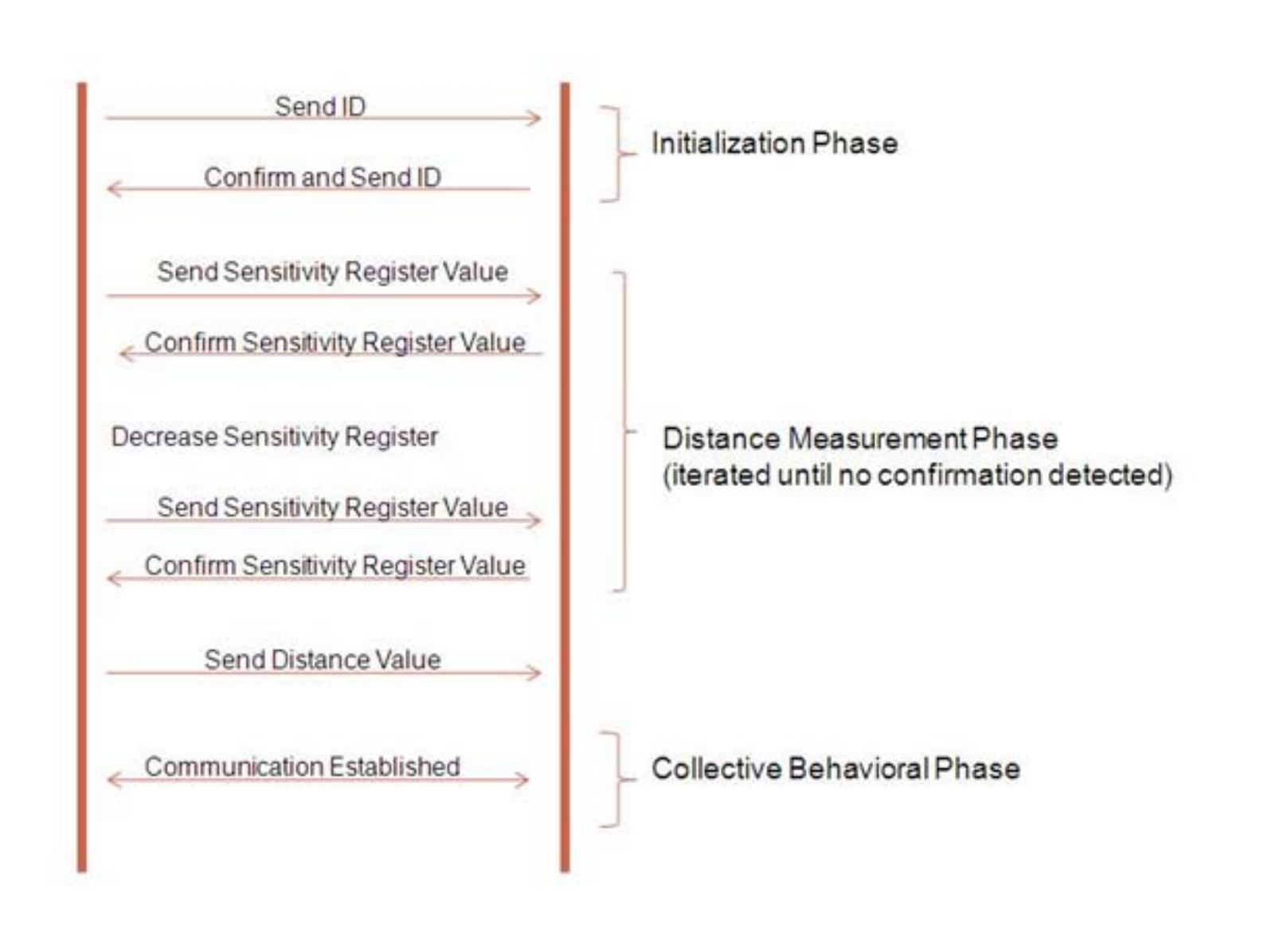}}
\caption{\textbf{(a)} Distance measurement with QAM blue light; \textbf{(b)} Active Sensing Algorithm.
\label{fig:chat_blue_light}}
\vspace{-2mm}
\end{figure}

The system must be configured and calibrated for finding the best modulation type for underwater communication and sensing. According to the measurement results, both for communication and sensing, the Quadrature Amplitude Modulation (QAM) seems to be the best modulation for underwater application. Fig.~\ref{fig:chat_blue_light}(a)
shows the relation between current sensitivity and communication distance. By using this curve, an active sensing algorithm can be added to the inter-robot communication algorithm by varying the amplification and the sensitivity of the programmable amplifier via software, see Fig.~\ref{fig:chat_blue_light}(b). The robot can approximate the distance with other robots by gradually decreasing the current sensitivity while communicating each other. The developed inter-robot communication algorithm has three phases. First, when two robots are in the communication range, they begin to establish the communication by sending their IDs to each other. Second, the communicating robots are approximating their distance by gradually decreasing the current sensitivity within the programmable gain amplifier. Therefore, after knowing their own position, behavioral or cooperation phased can be performed. A robot will continuously iterate the first phase if there is no other robots in the communication range. Hence an obstacle might reflect the transmitted signal and the robot would receive back the first phase communication packet that contains its own ID.

\subsection{Electric Sense}
\label{sec:elSense}

After experimenting with optical S\&C systems, we implemented another approach, which is inspired by weakly electric fish. These animals are capable of producing an electric field which they can use for localisation and communication~\cite{ANGELS}. Here we try to use this bio-inspired approach for analog communication and navigation in robot swarms.

\textbf{Electric fields.} Electric charges generate electrical fields in their vicinity. Electric fields are vector fields. For a point charge $Q$ the field intensity $\vec{E}$ can be calculated at each point $\vec{r}$ as
\begin{equation}
\label{eq2}
\vec{E} = \frac{Q}{4 \pi \epsilon_0 \epsilon_r} \cdot \frac{\vec{r}}{r^3}
\end{equation}
with the permittivities $\epsilon_0$ (vacuum) and $\epsilon_r$ (relativ) \cite{oatley76}.

The field vectors of multiple point charges follow the superposition principle. In our robot the electric field is generated by a dipole. The field intensity is proportional to the charge in the electrodes, which themselves are proportional to the applied voltage to the electrodes:
\begin{equation}
\label{eq3}
Q = C \cdot U
\end{equation}
with the voltage $U$ and the capacity $C$.

\textbf{Communication.} The intensity of an electric field can then be detected by measuring the differential potential between two electrodes in the field. This potential is proportional to the electric field intensity, which itself is proportional to the output voltage of the sender. By modulating the output voltage of the sender, information can be transmitted.

\textbf{Localization.} By using multiple pairs of electrodes in the receiver it is possible to calculate the bearing and distance to the sender. This is achieved by utilizing the drop in field intensity with relation to the distance. A sinus wave is impressed on the sender's electrodes which creates an oscillating electrical field. The field intensity depends mainly on the amplitude and frequency of the output voltage and some environmental conditions. The amplitude in the field intensity at a specific point $\vec{r}$ is proportional to the amplitude of the output voltage:
\begin{equation}
u(t) = a_o \cdot \sin(\omega t)   \sim   E(t)   \sim   a_i \cdot \sin(\omega t) \cdot \frac{\vec{r}}{r^3}
\label{eq:amplitudesim}
\end{equation}
with the amplitude of the output signal $a_o$ and measured input $a_i$, the frequency $\omega$ and the time $t$.

\begin{figure}[htp]
	\centering
\vspace{-2mm}
		\includegraphics[width=0.23\textwidth]{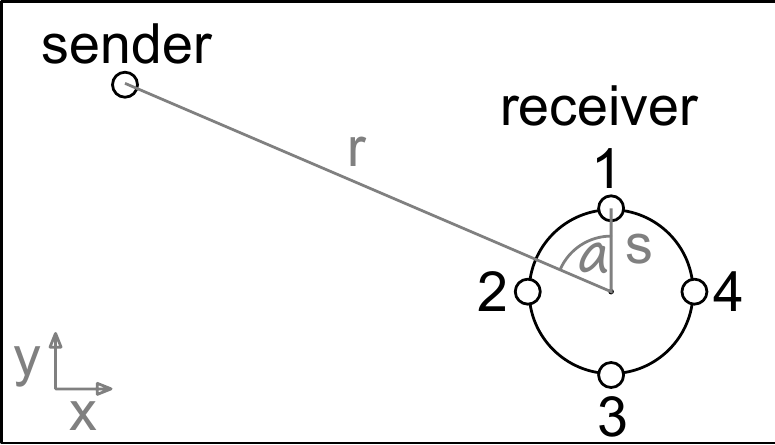}
		\includegraphics[width=0.23\textwidth]{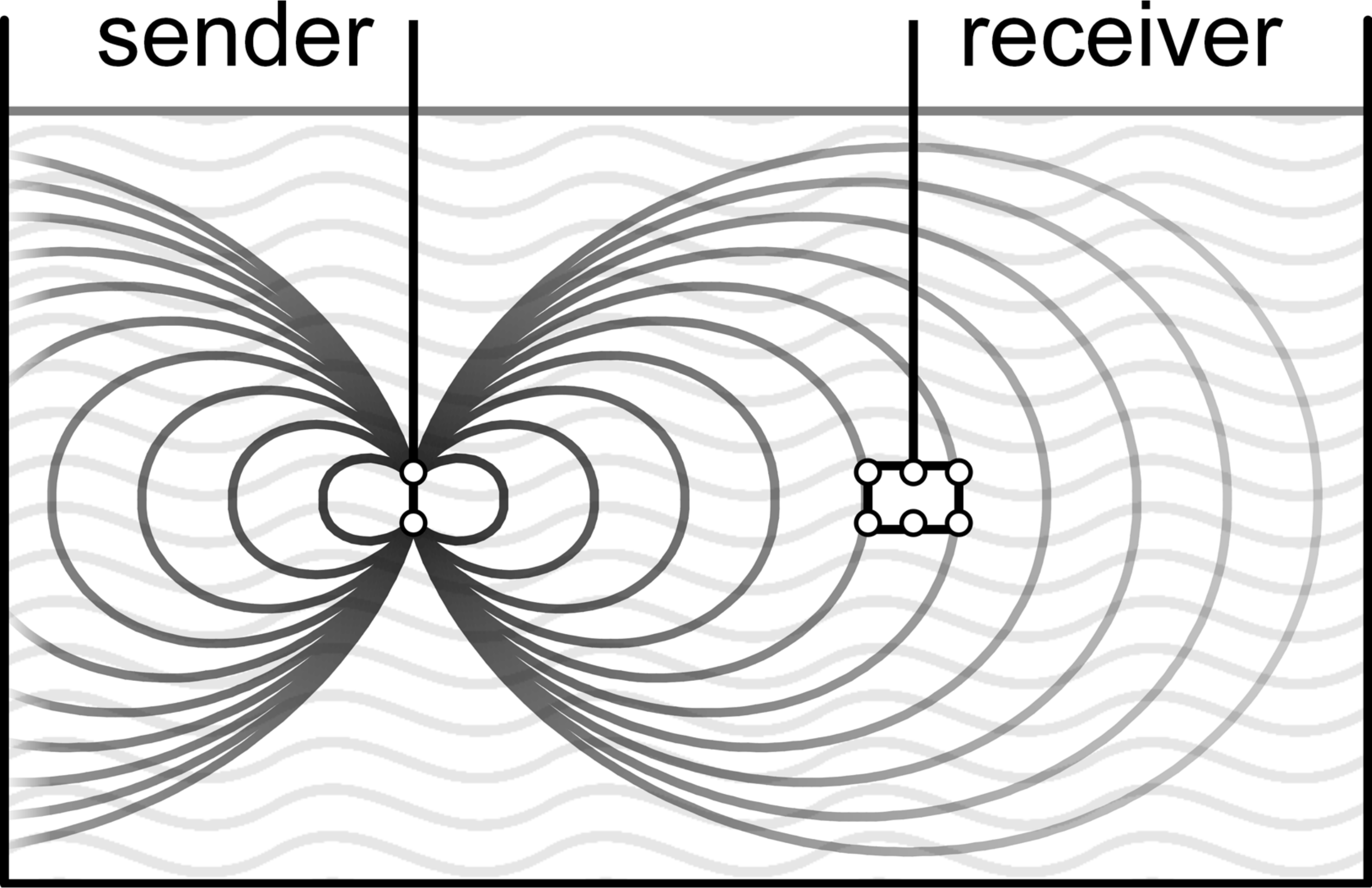}
	\caption{Position and orientation of sender and receiver electrodes, top-(left) and side-view (right)}
	\label{fig:versuch}
\vspace{-2mm}
\end{figure}

If sender and receiver are approximately in the same plane and the electrodes have the same orientation (compare Fig. \ref{fig:versuch} left) (\ref{eq:amplitudesim}) can be simplified to:
\begin{equation}
\label{eq4}
a_o \cdot \sin(\omega t) = a_i \cdot \sin(\omega t) \cdot \frac{F(\omega)}{r^2}
\end{equation}
with the frequency dependent proportionality factor $F(\omega)$ and the distance $r$ between sender and receiver. Measuring the sinus amplitude ($a_1$, $a_2$, $a_3$, $a_4$) at four points with a specific geometrical pattern (Fig. \ref{fig:versuch} right) leads to the following equations:
\begin{equation}
\begin{array}{r l r l}
a_1 = & \frac{A_o}{F(\omega)} \cdot \frac{1}{(r - s \cdot \cos \alpha)^2}, & a_2 = & \frac{A_o}{F(\omega)} \cdot \frac{1}{(r - s \cdot \sin \alpha)^2} \\
a_3 = & \frac{A_o}{F(\omega)} \cdot \frac{1}{(r + s \cdot \cos \alpha)^2}, & a_4 = & \frac{A_o}{F(\omega)} \cdot \frac{1}{(r + s \cdot \sin \alpha)^2} \\
\end{array}
\label{eq:amplitudessystem}
\end{equation}

In setting up these equations it is assumed that $r >> s$ so that the error in the angle $\alpha$ and distance $r$ between the different sensors is minimal. In (\ref{eq:amplitudessystem}) the proportional factor and output amplitude can be eliminated, under the condition of $r > s$ leading to:
\begin{equation}
r =  s \cdot \cos \alpha \underbrace{\frac{\sqrt{a_1/a_3} + 1}{\sqrt{a_1/a_3} - 1}}_{u_1},
r =  s \cdot \sin \alpha \underbrace{\frac{\sqrt{a_2/a_4} + 1}{\sqrt{a_2/a_4} - 1}}_{u_2}
\end{equation}
and
\begin{equation}
\label{eq6}
\alpha = \arctan \frac{u_1}{u_2}
\end{equation}

\textbf{Design limitations.} This approach has two design limitations: it requires the sender and receiver electrodes to have the same orientation (i.e. vertical) and to be roughly in the same plane (horizontal to the orientation):

\begin{itemize}
	\item The first limitation holds no practical difficulties. Our robot maintains a specific orientation, caused by its center of gravity. By placing one of the sender and receiver electrodes on top and one on the bottom of the robot the orientation is always vertical.
	
	\item The derivation above is only correct if sender and receiver are on the same plane, which is horizontal to the orientation of their electrodes. In a three dimensional environment this is not always true, but usually the working space is wider than it is high, even in a 3D environment.
\end{itemize}

Even so, we are working on overcoming these limitations. We are confident that the second can be eliminated by rearranging the receiver electrodes. To overcome the first limitation additional electrodes may be needed.

\section{Experiments}
\label{sec:experiments}

As described in the previous sections, different local S\&C systems utilize the same hardware components for sensing and communication. Moreover, they use modal and sub-modal approaches, which provide not only message transmission, but also deliver spatial information about position and distances of robots and objects. In this section we describe several behavioral experiments, performed with these systems.

\subsection{Experiments with Bi-modal Optical System}

\begin{figure}[htp]
\centering
\vspace{-2mm}
\subfigure{\includegraphics[width=.114\textwidth]{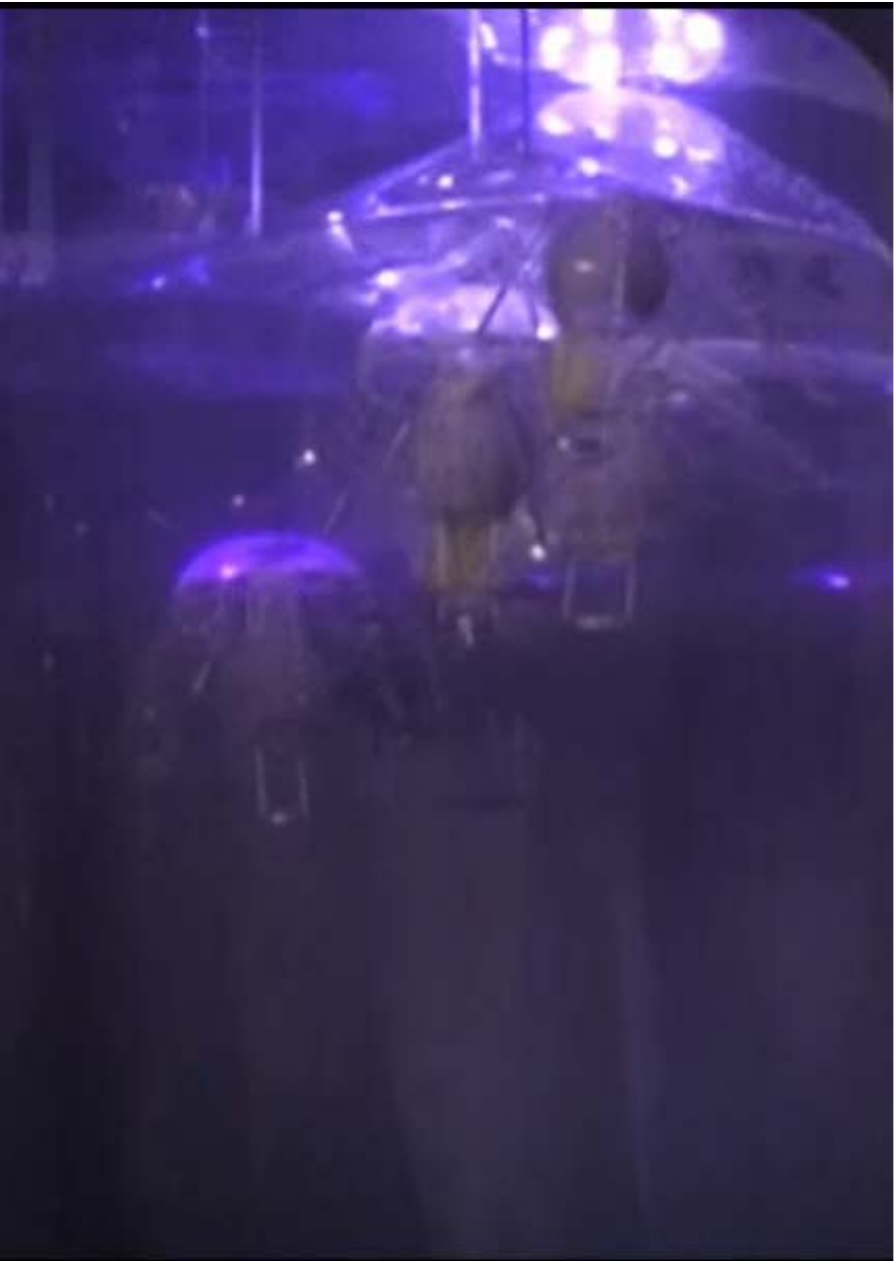}}
\subfigure{\includegraphics[width=.117\textwidth]{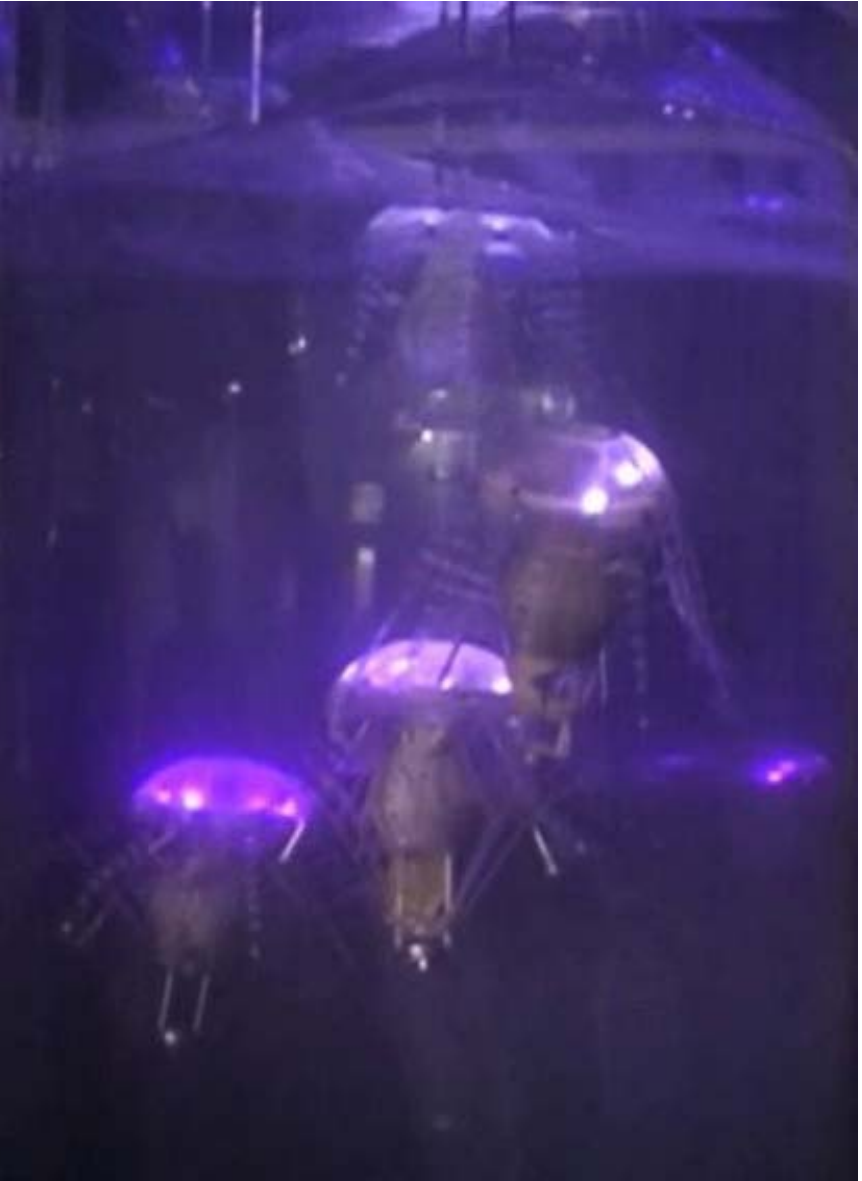}}
\subfigure{\includegraphics[width=.117\textwidth]{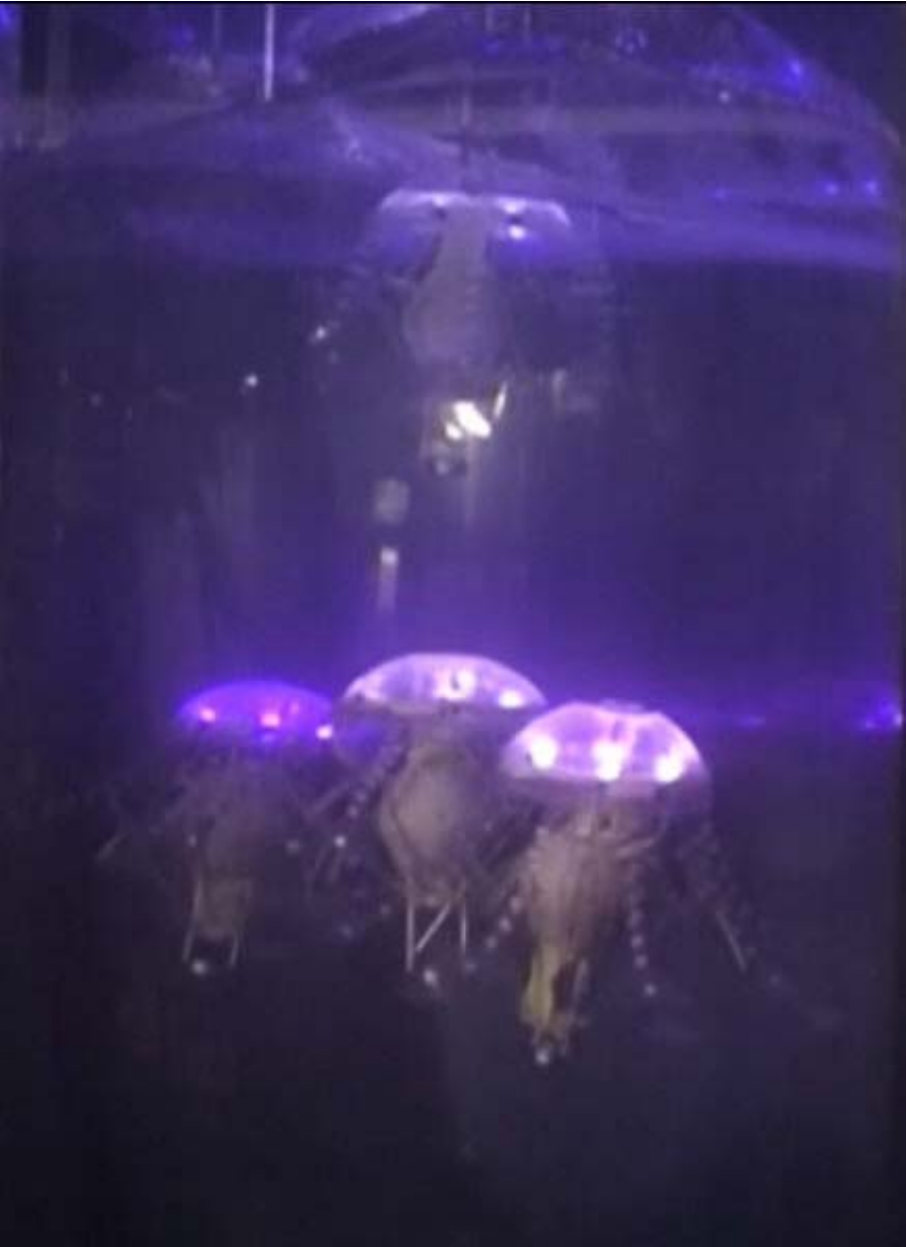}}
\subfigure{\includegraphics[width=.106\textwidth]{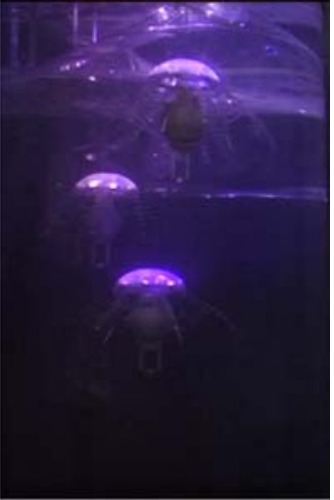}}
\caption{Experiment with collective decision making during the docking approach.\label{fig:scenario}}
\vspace{-2mm}
\end{figure}

One of the implemented scenarios with AquaJelly robots had the following form, see Fig.~\ref{fig:scenario}. In water, a robot sends sequentially in all IR channels its own ID. Listening and sending times are selected as approx. 95\% listening and 5\% sending, so that all robots most of time silently observe the environment. Receiving another-than-own ID means meeting another robot, whereas non-ID IR light means own reflection from passive objects. Granularity of IR channels is enough for rough collision avoidance with objects, e.g. walls of the aquarium. Collision avoidance based on digital channels are impossible for more than two robots (or robot and object).
In opposite, blue light channels emit almost all time. Since light has additive properties, when two robots meet each other, intensity of light in the point of light-spheres-intersection creates a light gradient and can be locally sensed by both robots. Especially interesting is the light gradient when several robots meet each other; they create complex gradients, which can be used for precise multi-robot navigation. Unfortunately, blue-light sensors continuously receive signals from their own light sphere so that no efficient communication is possible in this mode.

Initially all robots are fully charged. When a robot has a low energy value, it swims up and recharges. With the progress of experiment, more and more robots swim up for recharging. In this way, several robots meet in the upper part of the aquarium and compete for the docking station, see Fig.~\ref{fig:scenario}(from left to right). Since only a robot with lowest energy value should recharge, all robots bilaterally exchange values of their own energy level. The robot with the lowest energy value can swim up. This local behavior leads to the following interesting collective behavior. Due to light gradient created by many robots, all robots exert ``optical pressure" on each other, and collectively swim down, whereas only one most "hungry" robot swims up and recharges, see Fig.~\ref{fig:scenario}(right).

\subsection{Experiment with Encoded Blue Light}
\label{sec:elect}

In order to investigate the modulated blue light S\&C system, we used underwater submarine toy as a mechanical platform with new electronic components for locomotion, computational  and S\&C capabilities. This submarine has three degrees of freedom and three actuators for moving forward/backward, turning left/right, and diving up to 1 meter. The necessary modifications of the submarine including the replacement of the original electronic parts with the new designed electronic boards, and drilling some new holes on the robot's body for communication/sensing transducers placements. Cortex3 LM3S316 microcontroller with 25MHz of clock frequency, 16kb of internal flash ROM, and 16kb of RAM has been used in the platform. Two motor drivers and two navigational sensors are placed on the main board of the electronic platform. The combination of the available PWM output from Cortex3 and motor driver perform the ability to control the swimming velocity via software. A digital compass and pressure sensor are added as a three-dimensional orientation sensor (an low-frequency RF part is foreseen for a backup communication with host).
\begin{figure}[ht]
\centering
\vspace{-2mm}
\subfigure{\includegraphics[width=.24\textwidth]{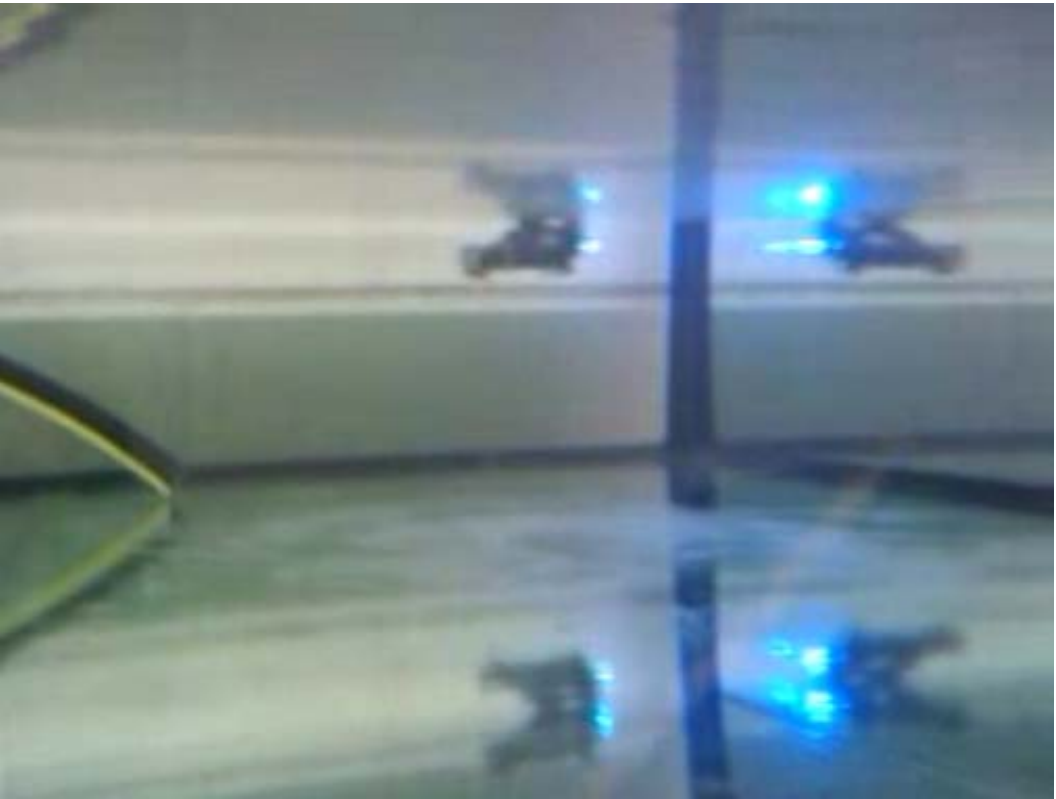}}~
\subfigure{\includegraphics[width=.24\textwidth]{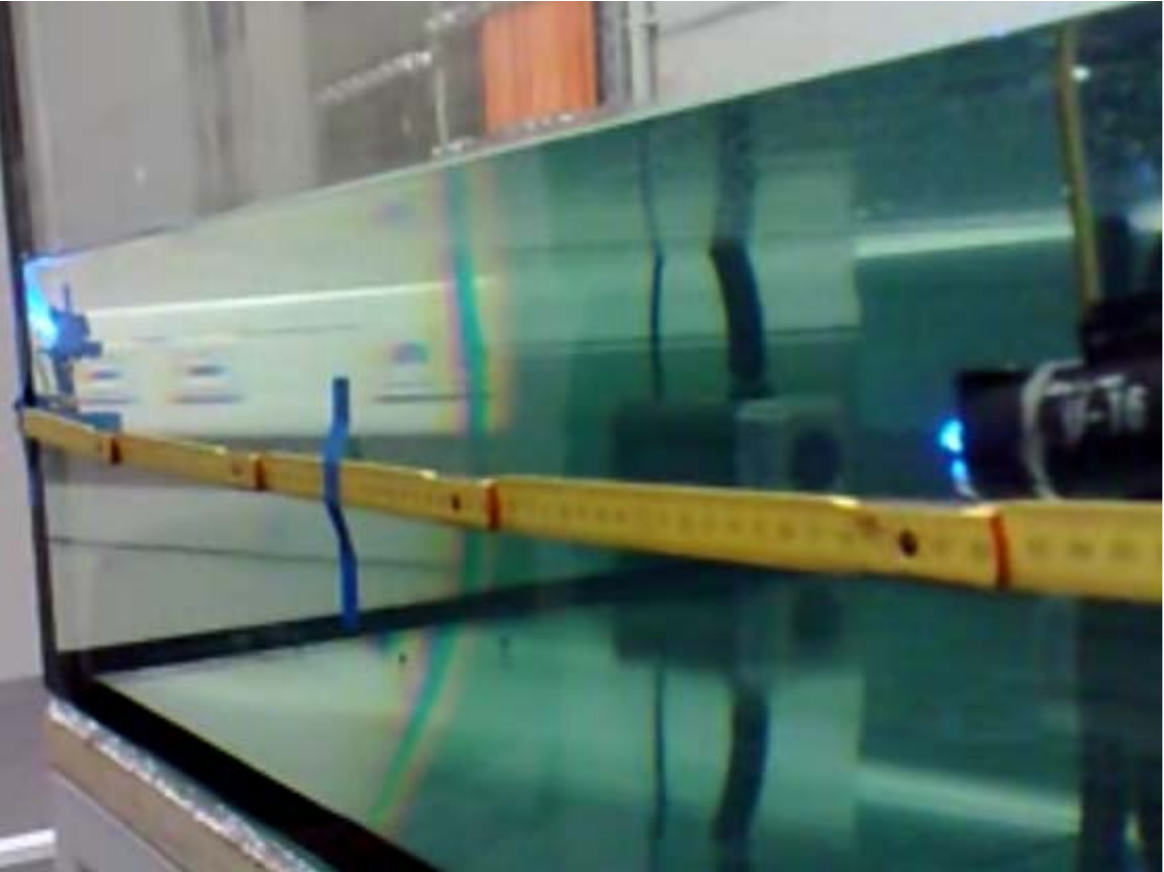}}
\caption{\textbf{(Left)} Autonomously swimming AUV recognizes obstacles with the digital S\&C system (active sensing); \textbf{(Right)} Experiments with digital communication. \label{fig:blue_light_platform}}
\vspace{-2mm}
\end{figure}

During the experiment, robot is deployed into the aquarium fulfilled with several obstacles. The available obstacles and aquarium's walls are used to examine the sensing capability, see Fig.~\ref{fig:blue_light_platform}(left). Both types of obstacles have different type of optical characteristic, which create different reflection behavior for the blue light. Therefore, white papers can be put outside the aquarium walls to increase the reflection capability of the optical sensor.

For testing the communication and active sensing capabilities, one robot is deployed underwater and a static encoded blue light transceiver is installed on the aquarium wall as a measurement reference point, see Fig.~\ref{fig:blue_light_platform}(right). The static transceiver can illuminate several type of light signals if it receives a specific blue light packet data from the swimming robot. Different types of blinking signals are used to examine the functionality of the active sensing capability. This approach underlies several other experiments, where a few passive robots are identified by one active AUV as foraging targets.

\subsection{Experiment with Electric Field}

\begin{figure}[htbp]
\vspace{-2mm}
	\centering
		\includegraphics[width=0.15\textwidth]{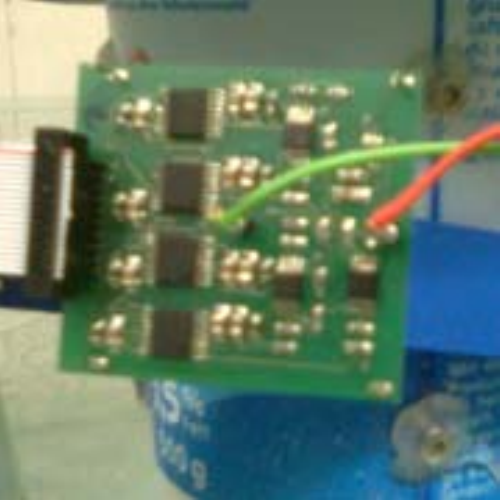}
		\includegraphics[width=0.15\textwidth]{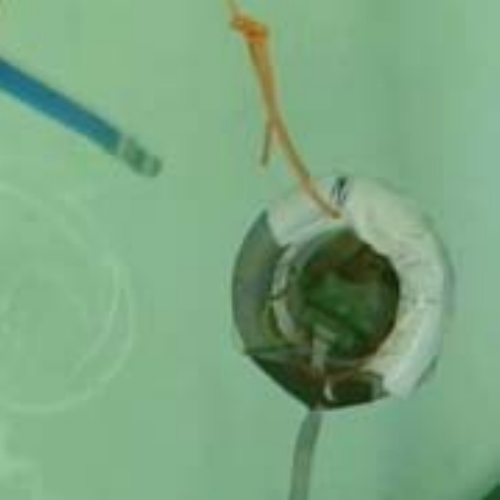}
		\includegraphics[width=0.15\textwidth]{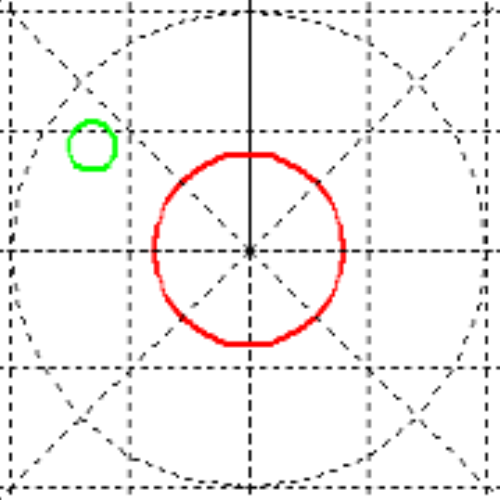}
	\caption{Experiment electric field: test circuit, video stream, calculated bearing}
	\label{fig:ESVersuch}
\vspace{-2mm}
\end{figure}

The circuit for electric field communication is very simple. It consists mainly of a digital-analog-converter (DAC) for the sender and four amplifiers (OP) with analog-digital-converters (ADC) for the receiver (Fig.~\ref{fig:ESVersuch}).

\begin{itemize}
	\item \emph{Sender:} The output of the 14-bit DAC is directly tied to one of the sender electrodes while the other is connected to VCC/2. The electrodes are in direct contact with the surrounding water. The output of the DAC can be set to a voltage between GND and VCC. This setup allows control of the field intensity and polarity.

	\item \emph{Receiver:} The receiver has four pairs of electrodes in the water to measure the difference in the potential of the electric field in four places. The electrodes use capacitors as highpass-filters to filter DC signals. The signals are amplified by differential OPs (magnitude 1000) and digitized by 14-bit ADCs.
\end{itemize}

For the experiments the sender and receiver are put under water (Fig.~\ref{fig:versuch} right). The receiver has a sampling rate of 10~kHz. The measured data is transmitted to a PC, where the bearing and distance are calculated.

\begin{figure}[htbp]
\vspace{-2mm}
	\centering
		\includegraphics[width=0.47\textwidth]{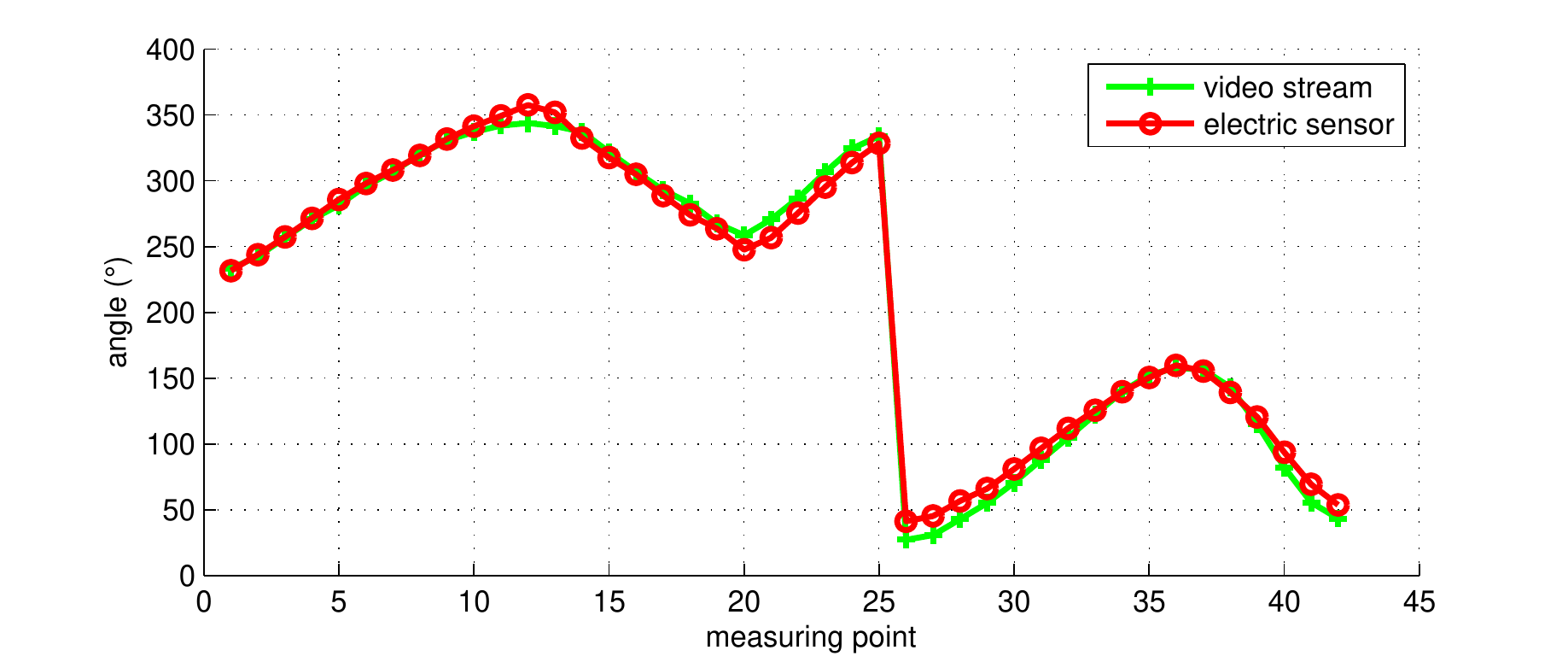}
	\caption{Calculated bearing from data provided by (a) video stream and (b) electrical sensor}
	\label{fig:vergleichcords}
\vspace{-1mm}
\end{figure}

In the experiment the sender was moved around the receiver and the received data was recorded together with a video tape of the experiment for comparison (Fig.~\ref{fig:ESVersuch}). The true bearing was extracted from the video stream and compared with the from the electrical sensor data calculated bearing. Fig.~\ref{fig:vergleichcords} shows the result.

For 50\% of the measuring points the error is less than 5$^{\circ}$ and it exceeds never more than 15$^{\circ}$. This might be further improved by increasing the magnitude of the amplifiers and reducing the noise through optimized circuits and digital filters.

\section{Conclusion}
\label{sec:conclusion}

In this work we considered several optical and electric-field-based approaches for sensing and communication within $R_c^l$. Together with S\&C approaches for $R_c^g$, such as acoustic and low-frequency RF, they represent the available spectra of S\&C technologies for underwater networked and swarm robotics. As indicated in the Sec.~\ref{sec:experiments} and from other performed experiments, these approaches combine communication with localization, distance measurement and object detection. In several cases, such a sub-modal information is available even during communication and can be used for very efficient behavioral strategies.

\begin{figure}[ht]
\centering
\vspace{-1mm}
\subfigure{\includegraphics[width=.4\textwidth]{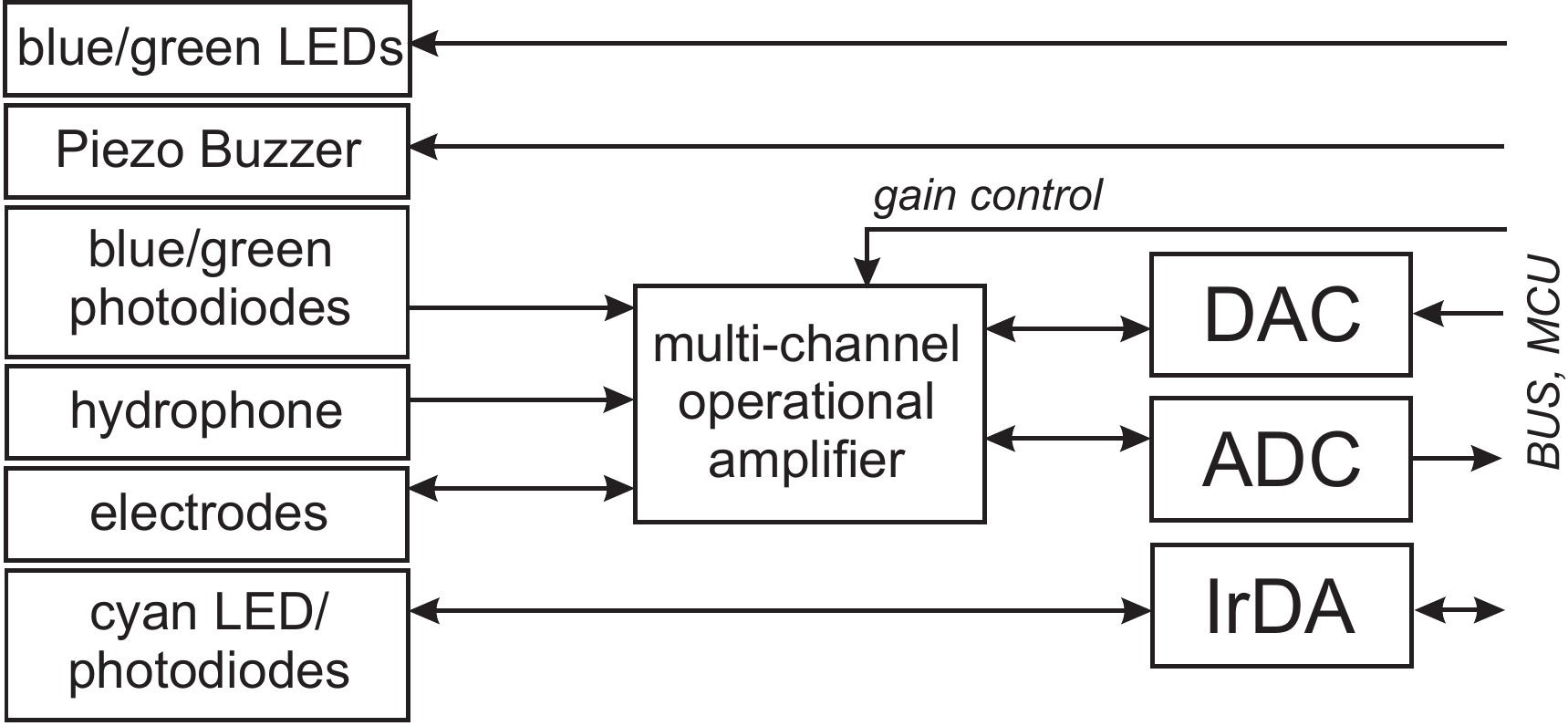}}
\caption{Combination of different S\&C approaches.
\label{fig:scheme}}
\vspace{-3mm}
\end{figure}

Each of the considered S\&C system has its own benefits and weaknesses. It seems that no current single system is capable of achieving all the requirements on $R_c^l$/$R_c^g$, sensing, minimal build space, energy consumption and complexity. Therefore the best approach lies in combining several of the available systems, for example in the way shown in Fig~\ref{fig:scheme}. 
The optical system provides split wave-length dependent channels. It can be used in analog and digital mode with existing control circuits for IR systems (e.g. IrDA or different modulations e.g. PCM/QAM), which with small modifications can be used for green, cyan and blue light. The range and bandwidth are sufficient for local communication. The channel is directional which can be of benefit for swarm-based coordination approaches. Additionally the reflection in analog mode can be used for navigation and detection tasks.

The electrical sensor is a good supplementary element to directional optics. Electric fields-based channels are omni-directional; hardware required for generation and detection of electric fields utilizes off-the-shelf components and is compact and energy efficient. It can be used to calculate the bearing between sender and receiver, i.e. for self-localization. The range is small but sufficient for $R_c^l$.

It is also necessary to supplement these S\&C systems by acoustic or ultra-low-frequency RF to provide global communication. Sonar requires a bit larger hardware equipment than the optical system. With additional components, it can be used for measuring distances to obstacles. RF systems represent a trade-off between the frequency (i.e. communication distances) and the size of integrated antennas (i.e. the size of platform). The control circuits are more complex than those for optic or acoustic approaches. Since bandwidth for low-frequency RF is not sufficient for application of standard protocols (e.g. ZigBee), global RF communication represents some open problems. If more than one receiver is used, the bearing between sender and receiver can be calculated. However this would make the hole system more complex and expensive. Comparing acoustic and low-frequency RF approaches for $R_c^g$, acoustic one is more favorable due to more less complex hardware. Usage of global communication for networked and swarm systems should be reduced to absolute minimum (see for instance minimalistic approaches for cooperation and decision making~\cite{Levi99}, \cite{Kornienko_OS01}).

\section*{Acknowledgment}

The ANGELS and CoCoRo projects are funded by the European Commission within the work programm ``Future and Emergent Technologies" and "Cognitive Systems and Robotics" under the grant agreements no. 231845 and 270382. We want to thank all members of the project for fruitful discussions.

\small
\IEEEtriggeratref{19}

\end{document}